\documentclass[
]{ceurart_preprint} 

\usepackage[frozencache=true,cachedir=minted-cache]{minted} 
\usepackage{subcaption}
\usepackage{listings} 
\usepackage[disable]{todonotes}
\usepackage{cleveref} 
\usepackage{hyperref} 

\usepackage[strict]{changepage}
\usepackage{framed} 
\usepackage{mathtools}

\usepackage{scrlayer-scrpage}
\clearpairofpagestyles
\definecolor{formalshade}{rgb}{0.95,0.95,1}
\definecolor{darkblue}{rgb}{0.0, 0.0, 0.55}

\newcounter{promptnumber}
\newenvironment{prompt}{%
 \refstepcounter{promptnumber}%
  \MakeFramed{\advance\hsize-\width\FrameRestore}%
  \noindent\hspace{-4.55pt}
  \begin{adjustwidth}{}{7pt}%
  \vspace{2pt}\vspace{2pt}%
   \textbf{Prompt \thepromptnumber:}~
}
{%
  \vspace{2pt}\end{adjustwidth}\endMakeFramed%
}

\Crefname{promptnumber}{Prompt}{Prompts}%
\crefname{promptnumber}{prompt}{prompts}%

\newcommand{\outlink}[1]{\allowbreak\, \raisebox{-0.7ex}{$\xrightarrow{#1}$} \, \allowbreak}
\newcommand{\inlink}[1]{\allowbreak\, \raisebox{-0.7ex}{$\xleftarrow{#1}$} \, \allowbreak}

\begin{document}

\copyrightyear{2023}
\copyrightclause{Copyright for this paper by its authors.
  Use permitted under Creative Commons License Attribution 4.0
  International (CC BY 4.0).}

\conference{dl4kg2023 @ ISWC: Workshop Deep Learning for Knowledge Graphs,
  November 6th-7th, 2023, Athen, Greece}

\title{Benchmarking the Abilities of Large Language Models for RDF Knowledge Graph Creation and Comprehension: How Well Do LLMs Speak Turtle?}

\author[1,2,3]{Johannes Frey}[orcid=0000-0003-3127-0815,email=frey@informatik.uni-leipzig.de]\fnmark[1]\cormark[1]
\author[1,2,3]{Lars-Peter Meyer}[orcid=0000-0001-5260-5181, email=lpmeyer@infai.org]\fnmark[1] \cormark[1]
\author[2,4]{Natanael Arndt}[orcid=0000-0002-8130-8677]
\author[1]{Felix Brei}[orcid=0009-0008-5245-6655]
\author[1,2]{Kirill Bulert}[orcid=0000-0002-1459-3754]

\address[1]{Institute for Applied Informatics, Goerdelerring 9, 04109 Leipzig, Germany, \url{https://infai.org}}
\address[2]{Agile Knowledge Engineering and Semantic Web (AKSW), \url{https://aksw.org}}
\address[3]{Leipzig University, Institute for Informatics, Germany, \url{https://www.uni-leipzig.de}}
\address[4]{eccenca GmbH, Leipzig, Germany, \url{https://eccenca.com}}

\cortext[1]{Corresponding author.}
\fntext[1]{These authors contributed equally.}

\begin{abstract}
Large Language Models (LLMs) are advancing at a rapid pace, with significant improvements at natural language processing and coding tasks. Yet, their ability to work with formal languages representing data, specifically within the realm of knowledge graph engineering, remains under-investigated.
To evaluate the proficiency of various LLMs, we created a set of five tasks that probe their ability to parse, understand, analyze, and create knowledge graphs serialized in Turtle syntax. These tasks, each embodying distinct degrees of complexity and being able to scale with the size of the problem, have been integrated into our automated evaluation system, the LLM-KG-Bench.
The evaluation encompassed four commercially available LLMs - GPT-3.5, GPT-4, Claude 1.3, and Claude 2.0, as well as two freely accessible offline models, GPT4All Vicuna and GPT4All Falcon 13B.
This analysis offers an in-depth understanding of the strengths and shortcomings of LLMs in relation to their application within RDF knowledge graph engineering workflows utilizing Turtle representation.
While our findings show that the latest commercial models outperform their forerunners in terms of proficiency with the Turtle language, they also reveal an apparent weakness. These models fall short when it comes to adhering strictly to the output formatting constraints, a crucial requirement in this context.
\end{abstract}


\maketitle

\section{Introduction}

Large Language Models have gained significant attention in recent years, with GPT-4 being among the most prominent \cite{Bubeck2023SparksArtificialGeneral}.
However, other models also demonstrate impressive performance in various tasks, as tracked in the LLMSYS Chatbot Arena Leaderboard\footnote{\url{https://huggingface.co/spaces/lmsys/chatbot-arena-leaderboard/tree/a068e66fdd6f453812b307541e8c82f99472aabe} \label{fn:leaderboard}}.

In the field of Knowledge Graph Engineering (KGE) the overarching task is to structure knowledge and encode it in a machine processable format.
Using machine learning techniques to create or process knowledge graphs is a well researched topic that receives new momentum.
The report of the Dagstuhl Seminar 22372 \cite{Groth2023Dagstuhl} and the Knowledge Base Construction from Pre-trained Language Models (LM-KBC) Challenge\footnote{Website: \url{https://lm-kbc.github.io/challenge2023/}} emphasize the relevance of this topic.
Pan et al. \cite{Pan2023UnifyingLargeLanguage} outline the potential of connecting LLMs and KGs and in particular with “LLM-augmented KGs”.

RDF - the Resource Description Framework - serves as a standard for representing Knowledge Graphs, while Turtle, a textual representation, is widely used to store and exchange RDF data. We have opted for Turtle given its strong resemblance to natural language, which aligns well with the primary mode of interaction with LLMs. 

In previous works, we have conducted manual experiments \cite{ours-AI-Tomorrow-23} and introduced the framework \emph{LLM-KG-Bench} \cite{ours-Semantics-23} for automated benchmarking of LLM performance on KGE tasks.

In this paper, we expand upon that work by introducing two new tasks to the \emph{LLM-KG-Bench} framework and evaluate the ability of various models  to shed light on the question “how well do LLMs speak Turtle”, i.e. parse, comprehend, analyze, create and serialize RDF knowledge graphs using Turtle serialization.
Besides Claude 1.3-100k, GPT-3.5 Turbo, and GPT-4, we also include Claude 2 in our evaluation and have extended the framework's capabilities to allow for benchmarking on a variety of freely available offline LLM models using GPT4All.
We have selected GPT4All Vicuna as a non-commercially usable model and Falcon 13B as the top freely available commercially usable model (Apache 2.0 license) to be additionally included. 

After describing related work in the next section, we introduce the benchmark tasks in \cref{sec:benchmark-tasks}. 
In \cref{sec:study}, we explain the study setup, present the evaluation and discuss the strengths and weaknesses of the individual LLMs regarding their utilization for RDF KGE workflows using Turtle.
We conclude with a discussion and outline future work in \cref{sec:conclusion}.

\section{Related Work}

There are several evaluations and articles discussing the utilization of LLMs for KG related tasks\footnote{Repository: \url{https://github.com/zjukg/KG-LLM-Papers/}}, e.g. \cite{Trajanoska2023EnhancingKnowledgeGraph, Carta2023IterativeZeroShot, zhu2023llms, Guo2023GPT4GraphCanLarge, ours-AI-Tomorrow-23}. 
Some of them include references to code for reproducing the results.
These works cover topics like, KG construction supported by LLMs, KG reasoning supported by LLMs or question answering based on KGs and LLMs.

In the field of generic LLM benchmarking, the \emph{BigBench} framework
\cite{Srivastava2022ImitationGameQuantifying} offers a robust structure and collects already a large list of automated benchmark tasks, but public code for integrating current LLMs like GPT or Claude
is missing.
The Large Model Systems (LMSys) leaderboard \cite{zheng2023judging}
is built mainly on manual testing and evaluation by the community.
There is also the Language Model Evaluation Harness \cite{lm-eval-harness} which tests Open Source LLMs on a variety of reasoning and logic tasks, but none are related to knowledge graphs.

In the scope of this paper, we are focusing on the definition of automated evaluation tasks testing KGE related capabilities using turtle syntax for RDF-based KGs.
We decided to use the KGE specific \emph{LLM-KG-Bench} framework\cite{ours-Semantics-23}, which is compatible with \emph{BigBench}, but adds KGE specific helpers, includes connectors for current LLMs like Claude and GPT, and supports  tasks that can scale in problem sizes.

\section{Benchmark Tasks}
\label{sec:benchmark-tasks}

To evaluate the capabilities of LLMs, we created five tasks with focus on the ability to parse, understand, analyze, and create knowledge graphs using the Turtle serialization format.

The tasks \textbf{T2} \emph{TurtleErrorsStatic} (\cref{sec:TurtleErrorsStatic}), \textbf{T3} \emph{TurtleSampleGeneration} (\cref{sec:TurtleSampleGeneration}), and \textbf{T5} \emph{FactExtractStatic} (\cref{sec:FactExtractStatic}) are extended versions of the tasks described in \cite{ours-Semantics-23}, while the tasks \textbf{T1} \emph{TurtleConnectionExplainStatic} (\cref{sec:TurtleConnectionExplainStatic}) and \textbf{T4} \emph{TurtleFriendCount} (\cref{sec:TurtleFriendCount}) are newly introduced in this paper.

The tasks are executed in two different manners, T1, T2, and T5 are executed as \textit{static} tasks, i.e. with a fixed prompt size and fixed expected responses, while T3 and T4 are \textit{scalable} in problem size (i.e. given prompt or expected response length) using an estimated byte limit parameter.
The byte limit can be used by the scalable tasks to calculate a task specific problem size (number of persons in the case of T3 and T4) to approximate that byte limit.

Task T1, T2 and T5 were executed 20 times per model. 
While the benchmark tasks report a variety of metrics and info or debug scores, we report the F1 measures for these tasks for a unified comparison in the scope of this work (shown in Figure \ref{fig:f1-measure-static-tests}).
The scalable tasks T3 and T4 were executed 20 times per combination of size and model for 8 different sizes.
Byte limit and resulting task problem sizes are depicted in table \ref{tab:scalableTask}.

\begin{center}
\captionof{table}{Configured byte limit and resulting task problem sizes} \label{tab:scalableTask}
\begin{tabular}{ c c c } \toprule[1.25pt]
 Byte Limit & No. Persons Task T3 & No. Persons Task T4 \\
 \hline
 1000 & 10 & 6 \\  
 2000 & 20 & 16 \\
 \vdots & \vdots & \vdots \\
 8000 & 80 & 76
\end{tabular}
\end{center}

\subsection{Task T1: Find Connection in Small Turtle File}
\label{sec:TurtleConnectionExplainStatic}
To check basic support for knowledge graphs and Turtle syntax, we implemented the \emph{TurtleConnectionExplainStatic} task similar to the first manual experiment in our previous work \cite{ours-AI-Tomorrow-23}.

\begin{prompt} \label{prompt:TurtleConnectionExplain}
    For the following turtle find the shortest non trivial connection from Anne to Bob. Please summarize the connection with just a list of resource IRIs, one per line, starting with https://abc.def/ghi/anne and ending with https://abc.def/ghi/bob . Please leave out rdf:type infos, leave out explanatory text and answer with just the IRI lines.
    
    [… followed by the graph in \cref{lst:organizational-kg}]
\end{prompt}

\begin{listing*}
\footnotesize
\begin{minted}[autogobble, linenos]{Turtle}
PREFIX : <https://abc.def/ghi/>
PREFIX rdfs: <http://www.w3.org/2000/01/rdf-schema#>
PREFIX owl: <http://www.w3.org/2002/07/owl#>
PREFIX foaf: <http://xmlns.com/foaf/0.1/>
PREFIX vcard: <http://www.w3.org/2006/vcard/ns#>
PREFIX org: <http://www.w3.org/ns/org#>

:anne a foaf:Person ; foaf:firstName "Anne" ; foaf:surname "Miller" ;
  vcard:hasAddress [ a vcard:Home ; vcard:country-name "UK" ] .
:bob a foaf:Person ; foaf:firstName "Bob" ; foaf:surname "Tanner" ;
  vcard:hasAddress [ a vcard:Home ; vcard:country-name "US" ] .

:wonderOrg a org:Organization .
:researchDep a org:OrganizationalUnit ; org:unitOf :wonderOrg ;
  rdfs:label "Research Department" .
:marketingDep a org:OrganizationalUnit ; org:unitOf :wonderOrg ;
  rdfs:label "Marketing Department" .

:chiefResearchOfficer a org:Role . :marketingManager a org:Role .

[ a org:Membership ; org:member :anne ; org:organization :researchDep ;
  org:role :chiefResearchOfficer ] .
[ a org:Membership ; org:member :bob  ; org:organization :marketingDep ;
  org:role :marketingManager ] .
\end{minted}
\caption{An organizational KG with two people working in different departments of the same organization. Graph taken from \cite{ours-AI-Tomorrow-23}.}
\label{lst:organizational-kg}
\end{listing*}

In \cref{prompt:TurtleConnectionExplain}, we provide a small organizational graph (see \cref{lst:organizational-kg}) and ask for the shortest non trivial connection  
(excluding the one via foaf:Person type statement) 
between the two nodes \emph{:Anne} and \emph{:Bob}.
By finding the connection  
$anne \inlink{org:member} bnode1  
\outlink{org:organization} researchDep 
\outlink{org:unitOf} wonderOrg 
\inlink{org:unitOf} marketingDep 
\inlink{org:organization} bnode2
\outlink{org:member} bob $ 
between them, the LLM demonstrates basic graph handling capabilities.
Note, that we ask the response output to be a list of resource/node IRIs without any other text, to support the automated evaluation of the answer. 
This also excludes both blank nodes and leads to the list $anne, researchDep, wonderOrg, marketingDep, bob$.
We use similar output requirements in most tasks and argue that a strict adherence to task details and output format instructions is a necessary capability for using LLMs as part of a tool chain in KGE tasks or workflows.
The task computes recall, precision, and F1 measure for the list of IRIs mentioned in the model response with regard to the list of IRIs representing the nodes of the shortest path.

\subsection{Task T2: Find Errors in Small Turtle File}
\label{sec:TurtleErrorsStatic}

\begin{prompt} \label{prompt:TurtleErrorsStatic}
    Please check the following rdf turtle file for errors and answer with no text but just the corrected turtle. Try to stick to the original formatting.
    [… followed by the turtle document to check]
\end{prompt}

The \emph{TurtleErrorsStatic} task involves identifying and correcting syntax errors in a small Turtle file, which is based on the same organizational graph (\cref{lst:organizational-kg}) with minor modifications. 
The turtle file has a period missing at the end of line 9 and the first semicolon in line 16 was removed. 
Correcting the errors demonstrates the LLM's knowledge of Turtle grammar while also showing its ability to transform it into a proper form without altering existing facts and adhering strictly to the task requirements.
One of the scores calculated during evaluation is the F1 measure on parsable, normalized triples, comparing the LLM’s answer with the perfect answer.
In order to do so, the response is directly consumed by \verb|rdflib| in combination with an iterative parsing failure heuristic.
The heuristic removes invalid lines that are reported as source of a syntax error until the document is fully parsable or empty.

\subsection{Task T3: Create Sample Graphs}
\label{sec:TurtleSampleGeneration}
We created the task \emph{TurtleSampleGeneration} to see if LLMs can understand and honor the requirements that we postulate for creating a simple knowledge graph, i.e. the number of resources in the graph as well as its structure.
The task makes use of the popular \verb|FOAF| vocabulary because we assume that members of it are very prevalent in the training data, since they are frequently used in example snippets in online forums and datasets.

\begin{prompt} \label{prompt:TurtleSampleGeneration}
    Create a knowledge graph in turtle format that consists of $n$ different objects of type \verb|foaf:person|. Each should have at least 2 and at most $n-1$ connections to other persons via \verb|foaf:knows|. Give no extra text.
\end{prompt}

In \cref{prompt:TurtleSampleGeneration} we instruct the LLM to generate a knowledge graph with a certain number of persons who each have between two and $n-1$ friends (both inclusive). 
The number of persons $n$ can be varied to get different answer sizes. 
The task is motivated by the idea of using LLMs to generate test, training, or example data of various sizes.
Furthermore, it allows to study the capacity of the models to generate content of increasing sizes while maintaining the integrity of the graph and serialization.
This task checks first if the generated answer is parsable. If so, the structure of the graph is evaluated.
We count the number of resources in the graph and if they are all correctly declared as \verb|rdf:type| \verb|foaf:Person|. 
The \emph{persons\_relative\_error} scores measures the difference between the actual number of person objects generated and the number asked for.
This value is normalized to be $=0$ if they match, $>0$ if there are more persons than asked for and $<0$ if there are less persons, with the special case of $-1$ meaning an empty graph.

\subsection{Task T4: Count Links in Person Graph}
\label{sec:TurtleFriendCount}

\begin{prompt}
    \label{prompt:TurtleFriendCount}
    Please name the person which is known by the most persons according to the following RDF graph in turtle syntax. Give just the IRI of this person with most incoming links as answer, without abbreviation or explanation.
    [… followed by the graph serialization]
\end{prompt}

The task \emph{TurtleFriendCount} requires finding the resource with the most incoming links in a simple generated KG.
The structure of the KG is similar to the previous tasks, consisting of a variable number of \verb|foaf:Person| resources connected by \verb|foaf:knows| properties.
Each person is known by two other persons, but one designated \verb|foaf:Person| is known by two additional persons (one for small sizes), resulting in three to four incoming links instead of two.
This task tests basic RDF and turtle knowledge as well as graph comprehension and processing skills by aggregating link counts for various KG sizes.
The number of \verb|foaf:Person| resources in the prompt has a linear correlation with the prompt length.\todo{explain from char sizes to person number}
The task computes recall, precision, and f1 measure with respect to the expected person IRI.

\subsection{Task T5: Create Knowledge Graph from Factsheet}
\label{sec:FactExtractStatic}

\begin{prompt}
    \label{prompt:FactExtractStatic}
As a Linked Data expert and knowledge engineer please convert the 3d printer specification given in the bottom into an RDF turtle formatted Knowledge Graph.
The main subject in this Knowledge Graph should be the Printer itself using the https://data.semper-ki.org/resources/\$encoded\_label\$ whereas \$encoded\_label\$ label refers to the name of the printer in which whitespaces have been replaced with underscores. Step by step attach the following statements to that node.\\
1) Attach https://purl.org/tema/051993 as rdf:type\\
2) Attach the manufacturer via an object property using the schema.org vocabulary. For the subject identifier of the manufacturer use the same approach as for the printer subject and assign it the rdfs:label without a language tag.\\
3) Attach all printable material as well all support materials the printer can use via object properties. For the property identifiers use https://data.semper-ki.org/properties/printMaterial  respectively https://data.semper-ki.org/properties/supportMaterial and for material identifiers use the https://data.semper-ki.org/resources/materials/ namespace.\\
4) Attach the dimensions of the printer as 3 separate values using the schema.org vocabulary for the property identifiers of the dimensions, but for the values use qudt:QuantityValue objects from the QUDT ontologies family to have the numerical value ( via qudt:numericValue property and typed as xsd:double) and the appropriate QUDT unit of measurement identifier (via qudt:hasUnit property) separately. Do not convert the dimensions into another unit.\\
Follow best practices of Linked Data and common standard vocabularies as basis for the modeling unless specified differently in the requirements above.
Do not extract any other properties or values from the specification than the ones I mentioned but validate your output step by step, to check whether all used prefixes are included and the given requirements as well as the grammar of the RDF turtle serialization were respected. 
Only return the turtle format, no additional text.
 [... followed by the fact sheet plaintext excerpt]
\end{prompt}

The task \emph{FactExtractStatic} assesses the LLM's fact extraction and advanced RDF modeling abilities, by utilizing a plaintext excerpt from one of our previous experiments \cite{ours-AI-Tomorrow-23}.
The excerpt (that is not shown for reasons of brevity) describes various aspects of the specifications of a 3D printer in a key-value format, including the formatting irregularities commonly found in PDF extracts.
We ask the model to generate a Turtle file that captures a subset of this information to check for how well RDF facts can be extracted from factsheet plaintexts and transformed into a knowledge graph.
The prompt is carefully designed with regard to the transformation aspect in order to be very specific and unambiguous on how the data should be represented.
The prompt defines concrete namespace schemes and construction rules for IRI identifiers for properties and subjects, but also challenges knowledge about ontology members by requesting the use of concrete ontologies.
Subsequently, we can evaluate the quality of a single response using the F1 measure, counting the set of parsable triples that (mis)match or are missing compared to a manually curated reference document.
Since we consider this as a quite challenging task, we make use of multiple prompt engineering optimization techniques, namely, asking for an expert answer in a domain context, providing step by step instructions, and asking for critical self-evaluation.

\section{Benchmark Study Results and Discussion}
\label{sec:study}
Using the \emph{LLM-KG-Bench} framework, we configured the 5 aforementioned benchmark tasks (cf.~\cref{sec:benchmark-tasks}) to be evaluated for the 3 highest ranking LLMs at the LLMSYS Chatbot Arena Leaderboard\textsuperscript{\ref{fn:leaderboard}} i.e.~GPT-4 (gpt-4-0613), GPT-3.5 (gpt-3.5-turbo-0613), and Claude-1.3 (claude-1.3-100k), additionally we have included Claude-2.0.
These 4 systems were evaluated using the commercial APIs of OpenAI and Anthropic.
We also wanted to include freely available offline LLMs.
Based on the availability in GPT4All, we have selected GPT4All Vicuna-13B (version \texttt{ggml-vicuna-13b-1.1-q4\_2.bin}) as a non-commercially usable representant and GPT4All Falcon (\texttt{ggml-model-gpt4all-falcon-q4\_0.bin}) as the top freely available commercially usable model (Apache 2.0 license).

\begin{figure}
  \begin{subfigure}{0.33\textwidth}
    \centering
    \includegraphics[width=\linewidth]{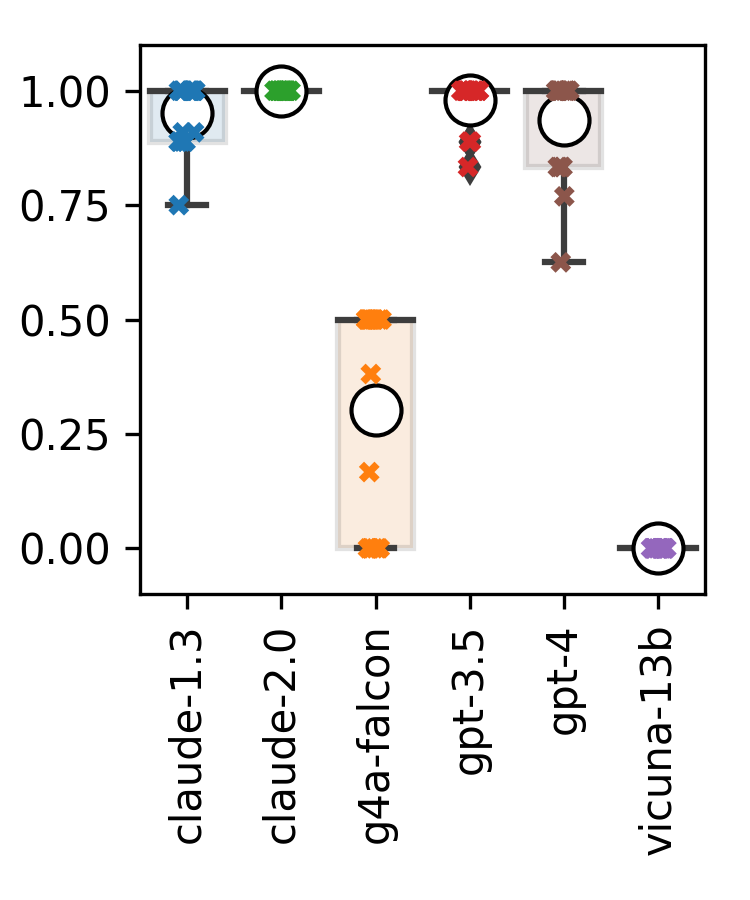}
    \caption{T1 - Connected Path}
    \label{fig:ShortestPath}
  \end{subfigure}
    \hfill
  \centering
  \begin{subfigure}{0.33\textwidth}
    \centering
    \includegraphics[width=\linewidth]{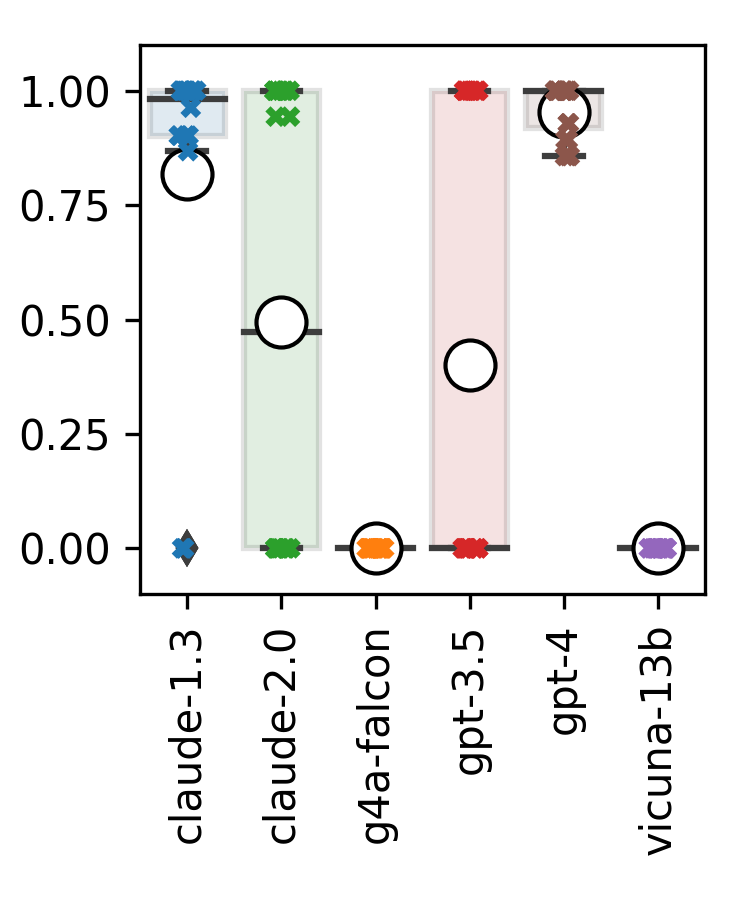}
    \caption{T2 - Turtle Fixing}
    \label{fig:TurtleErrorFixingScores}
  \end{subfigure}%
    \hfill
  \begin{subfigure}{0.33\textwidth}
    \centering
    \includegraphics[width=\linewidth]{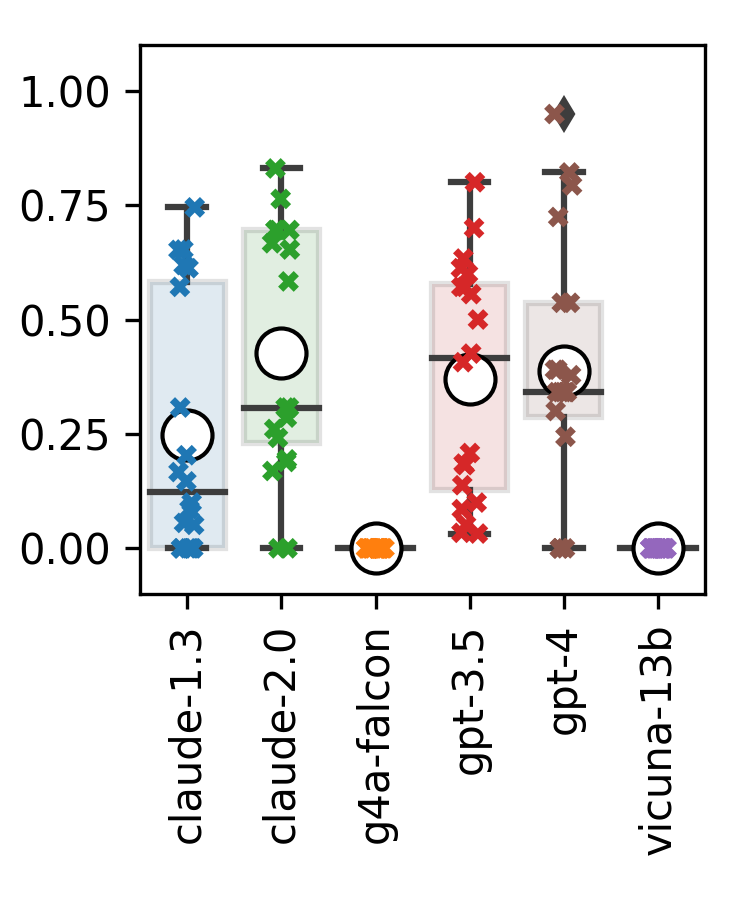}
    \caption{T5 - Fact Extraction}
    \label{fig:plotFactsExtract}
  \end{subfigure}%
  \caption{Evaluation of Static Tasks: Distribution of F1 scores}
  \label{fig:f1-measure-static-tests} 
\end{figure}

\begin{figure}
  \centering
    \begin{subfigure}{0.49\textwidth}
    \centering
    \includegraphics[width=\linewidth]{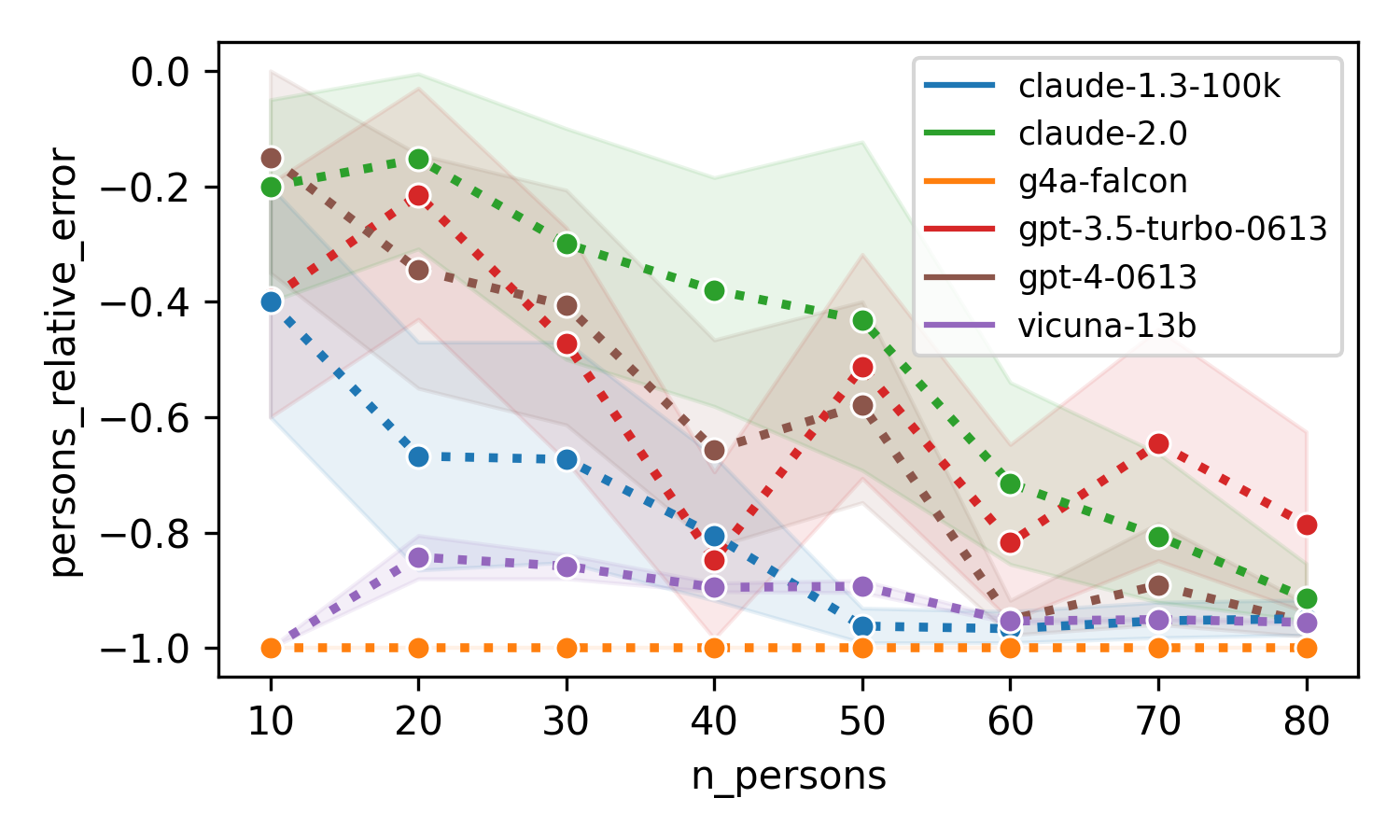}
    \caption{T3 - Person Graph Creation}
    \label{fig:PersonGraphGeneration}
  \end{subfigure}
  \hfill
    \begin{subfigure}{0.49\textwidth}
    \centering
    \includegraphics[width=\linewidth]{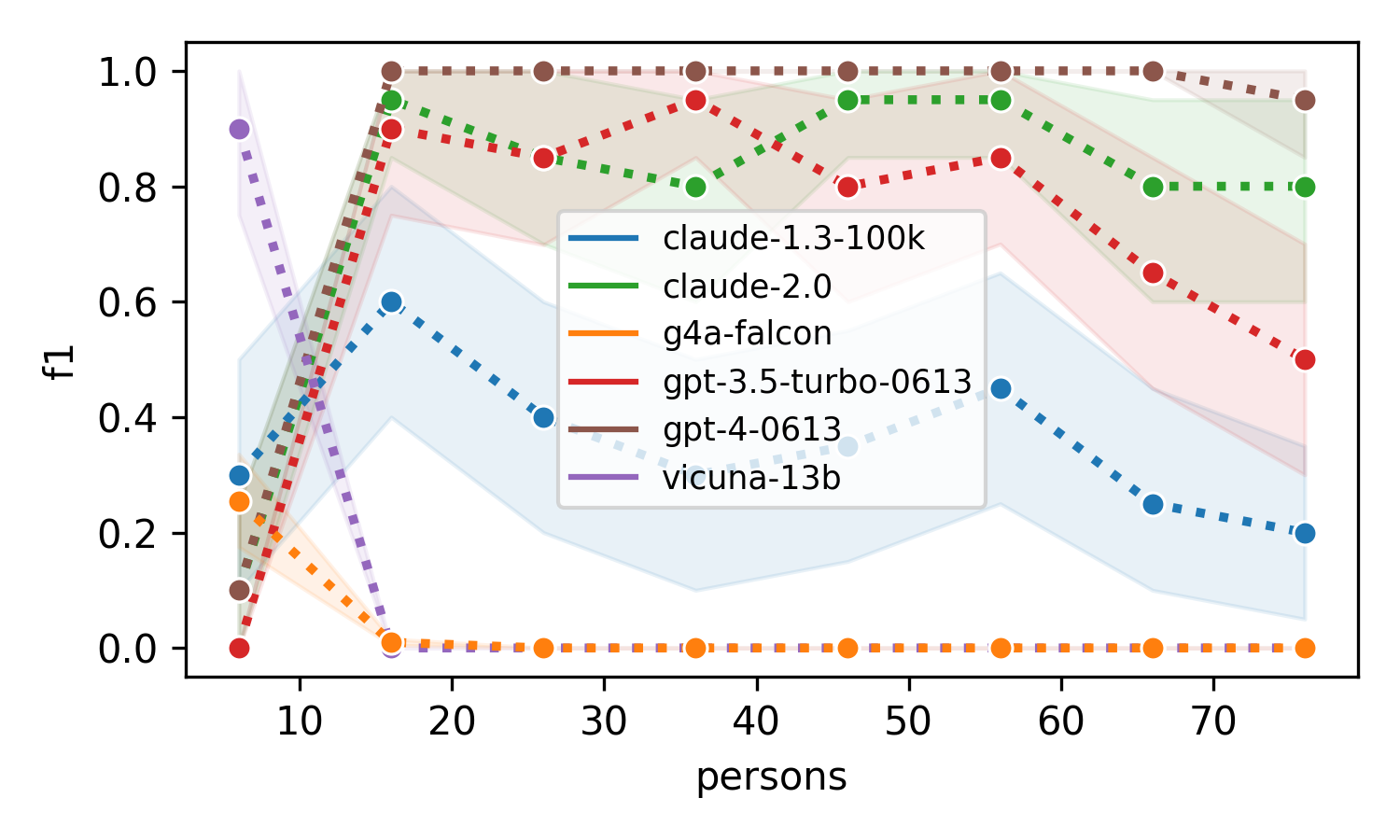}
    \caption{T4 - Friends Counts}
    \label{fig:ConnectionCounts}
  \end{subfigure}%
  \caption{Evaluation of Scalable Tasks: Mean of task metric with 95\% confidence interval}
  \label{fig:measures-scalable-tests} 
\end{figure}

\textbf{T1:}
As can be seen in Figure \ref{fig:ShortestPath}, Claude-2.0 answers perfectly.
GPT-4 added occasionally properties to the list, although only nodes were requested.
Claude-1.3 and GPT-3.5 seem to omit sometimes a resource from the requested list.
Falcon's lists often contain only the first and last resource which was already specified in the prompt, but a few attempts pose some basic understanding of the task, but severely violate the output constraints.
Vicuna`s responses are mostly in the form of "This is the shortest connection from anne to bob".

\textbf{T2:}
GPT-3.5 often claims that the given turtle file would be correct and returns no turtle. 
This explains the high rate of zeros as F1 score (see figure \ref{fig:TurtleErrorFixingScores})
The answers given by Claude-1.3 and GPT-4 score better.
Claude 2.0 fails in contrast to its predecessor in returning the plain turtle, leading to unparseable documents although the errors seem fixed.
Vicuna replies with the empty string in all cases. 
While Falcon reports there would be no errors in a few cases, it often does not follow the task and replies with Turtle snippets or explanations of the content.

\textbf{T3:}
Claude-1.3 clearly performed worst from the commercial models (Fig. \ref{fig:PersonGraphGeneration}). 
Common issues leading to missing persons for smaller sizes were omitting rdf:type statements and missing prefix definitions (making the entire document unparseable). 
For bigger sizes, Claude 1.3 additionally used ellipses with comments like "continues with 70+ more persons" making it impractical for sample generation.
Claude 2.0 shows similar ellipses but sovereignly handles types and prefixes, leading to the best results for small and medium sizes, but bad ones for large sizes.
GPT-4 has only a slightly better performance compared to GPT-3.5 but in contrast uses ellipses much more frequently for higher sizes (60-80 persons).
Falcon and Vicuna scored worse. 
Vicuna creates very few persons but then uses ellipses. 
However, it is missing the essential rdf:type statement for size 10.
Falcon just creates a list of prefix declarations.

\textbf{T4:}
All commercial models seem to be challenged by the size of 6 persons as indicated in Figure \ref{fig:ConnectionCounts}, where the person with the most friends only has one more friend compared to all others. 
GPT-3.5 consistently proposes the person with the most outgoing friends as solution instead. 
For the other models this also happens very frequently. 
Surprisingly, this potential misunderstanding seems to occur significantly less frequent for all other sizes where the the person with the most friends has 4 incoming relations.
In fact, GPT-4 is performing in an outstanding way, only doing one mistake for the largest size. 
Claude-2.0 performs similar to GPT-3.5, both confuse one of the persons with more outgoing links as correct solution, although interestingly GPT-3.5 also answers often with a full sentence (instead the IRI only) when it has such a confusion.
Claude 1.3 also frequently confuses ingoing with outgoing links, and similarly to GPT 3.5 this correlates with violating the output constraint, but even stronger.
Falcon and Vicuna seem to understand the task but have incorrect reasoning.
Falcon consistently answers with a wrong non-reasonable person and textual explanations.
Vicuna surprisingly identifies the correct person for the smallest size, but fails for bigger sizes. 
Moreover the context windows are exceeded for sizes greater than 26 persons for Vicuna and 36 persons for Falcon. 

\textbf{T5:}
Figure \ref{fig:plotFactsExtract} shows that both GPT models outperform Claude 1.3 in this task. 
While GPT4 has a better mean, due to one very good response (F1 score of 0.94), it however replied frequently with unparseable content, which in turn did not happen for GPT3.5, leading to a slightly better median for that. 
Claude 2.0 shows the highest values for F1 mean as well as the  third quartile and it returned less unparseable documents  compared to Claude 1.3 and GPT4.
Vicuna did not return output that has any similarity with Turtle.  
Falcon creates turtle preambles with varying prefix definitions but had problems to continue the document and seemed very often stuck in repetitive namespace gibberish and prefix patterns.

\section{Conclusion and Future Work}
\label{sec:conclusion}
The evaluation shows already promising results. 
Especially the newer versions of both GPT and Claude speak turtle already at a level that it might be useful for assistant tasks. 
A general problem is though, that the models, although explicitly requested, do not consistently respond with plain Turtle but include short explanations or markdown ticks. 
While our parsing failure heuristic can mitigate some of these cases, this issue poses a challenge when interfacing directly with RDF tools. 
It is noteworthy, that the newer versions of both GPT and Claude tend to violate the output constraints more often.
While our failure tolerant parser heuristic allows to get more insights into the quality of results, it can also reward solutions to some degree that might not be useful without special post processing in practical scenarios.
Therefore, we see as a next step to define tests that are stricter, however provide feedback to the models (e.g. parsing errors) to perform and evaluate few-shot approaches.
Moreover, it could be of value to assess the performance using ntriples, which has less syntactical features but allows easier retrieval of partially inconsistent responses.
It also remains to be seen whether finetuning LLMs on RDF syntax using large datasets like Wikidata and DBpedia would be beneficial. 
Finally besides extending the framework with more tests, we see integrating LangChain to study the combination of LLMs with KGE-assistant plugins (e.g. prefix or ontology terminology lookup service) as an interesting path to explore.

\begin{acknowledgments}
  This work was partially supported by grants from the German Federal Ministry of Education and Research (BMBF) to the projects StahlDigital (13XP5116B) and KupferDigital (F13XP5119F) as well as from the German Federal Ministry for Economic Affairs and Climate Action (BMWK) to the CoyPu project (01MK21007A) and KISS project (01MK22001A).
\end{acknowledgments}

\bibliography{ref}

\appendix

\section{Online Resources}

\begin{itemize}
  \item Source Code Repository: \url{https://github.com/AKSW/LLM-KG-Bench} or \href{https://doi.org/10.5281/zenodo.8366061}{doi:10.5281/zenodo.8366061}
  \item Study Data: \url{https://github.com/AKSW/LLM-KG-Bench-Results/tree/main/2023-DL4KG_Turtle-KG-Eval} or \href{https://doi.org/10.5281/zenodo.8364535}{doi:10.5281/zenodo.8364535}
\end{itemize}

\end{document}